\documentclass[final]{cvpr}

\usepackage{times}
\usepackage{epsfig}
\usepackage{graphicx}
\usepackage{amsmath}
\usepackage{amssymb}

\usepackage[pagebackref=true,breaklinks=true,colorlinks,bookmarks=false]{hyperref}

\usepackage{amssymb} 

\usepackage{enumitem}
\usepackage{booktabs}

\usepackage{amsmath}
\usepackage{xspace}

\usepackage[dvipsnames,table]{xcolor}
\usepackage[pagebackref=true,breaklinks=true,colorlinks,bookmarks=false]{hyperref}

\usepackage{cleveref}
\definecolor{darkergreen}{RGB}{21, 152, 56}
\definecolor{red2}{RGB}{252, 54, 65}
\newcommand\redp[1]{\textcolor{red2}{(#1)}}
\newcommand\greenp[1]{\textcolor{darkergreen}{(#1)}}

\definecolor{lightblue}{rgb}{0,0,1}

\usepackage{arydshln}

\usepackage{pifont}
\usepackage[table]{xcolor}
\usepackage{algorithm}
\usepackage{algorithmic}

\newcommand*\colorcmark[1]{%
  \expandafter\newcommand\csname #1cmark\endcsname{\textcolor{#1}{\ding{51}}}%
}
\colorcmark{green}

\newcommand*\colorxmark[1]{%
  \expandafter\newcommand\csname #1xmark\endcsname{\textcolor{#1}{\ding{55}}}%
}
\colorxmark{red}

\usepackage{multirow}
\usepackage{enumitem}
\usepackage{subcaption}
\usepackage[table]{xcolor}
\usepackage{amsthm}
\newlength\savewidth\newcommand\shline{\noalign{\global\savewidth\arrayrulewidth
  \global\arrayrulewidth 1pt}\hline\noalign{\global\arrayrulewidth\savewidth}}

    \newlength\thinwidth
    \definecolor{Gray}{gray}{0.92}
    \newcolumntype{a}{>{\columncolor{Gray}}c}
    \definecolor{LightCyan}{rgb}{0.88,1,1}
    \definecolor{lightblue}{rgb}{0,0,1}
    \definecolor{darkergreen}{RGB}{21, 152, 56}
    \definecolor{highlightRowColor}{gray}{0.92}

\def\cvprPaperID{4008} 
\def\confYear{CVPR 2021}

\begin{document}

\title{Removing the Background by Adding the Background: Towards Background Robust Self-supervised Video Representation Learning}

\author{Jinpeng Wang$^{1,2}$ \thanks{The first two authors contributed equally. This work was done when Jinpeng was in Tencent Youtu Lab.} \quad Yuting Gao$^{2}\;$\footnotemark[1] \quad Ke Li$^2$ \quad Yiqi Lin$^1$ \quad Andy J. Ma $^1$ \thanks{Corresponding Author}\\ Hao Cheng $^2$ \quad Pai Peng $^{2}$ \quad Feiyue Huang $^{2}$ \quad Rongrong Ji$^3$ \quad Xing Sun$^2$\\ \\
$^1$Sun Yat-sen University\quad $^2$Tencent Youtu Lab\quad $^3$Xia Men University\\}

\maketitle


\begin{abstract}
Self-supervised learning has shown great potentials in improving the video representation ability of deep neural networks by getting supervision from the data itself. However, some of the current methods tend to cheat from the background, \emph{i.e.}, the prediction is highly dependent on the video background instead of the motion, making the model vulnerable to background changes. To mitigate the model reliance towards the background, we propose to \emph{remove the background impact by adding the background}. That is, given a video, we randomly select a static frame and add it to every other frames to construct a distracting video sample. Then we force the model to pull the feature of the distracting video and the feature of the original video closer, so that the model is explicitly restricted to resist the background influence, focusing more on the motion changes. We term our method as \emph{Background Erasing} (BE). 
It is worth noting that the implementation of our method is so simple and neat and can be added to most of the SOTA methods without much efforts. Specifically, BE brings 16.4\% and 19.1\% improvements with MoCo on the severely biased datasets UCF101 and HMDB51, and 14.5\% improvement on the less biased dataset Diving48.

\end{abstract}

\section{Introduction}

\begin{figure}[hbt!]
	\centering
	\includegraphics[width=\linewidth]{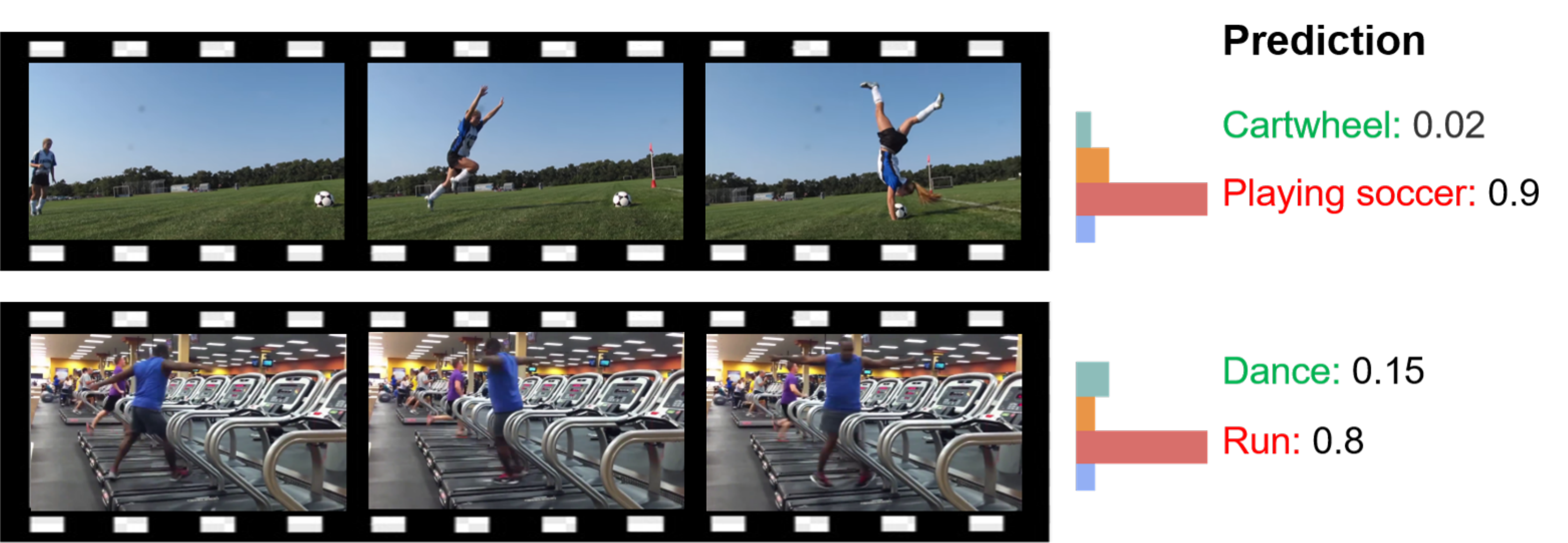}
	\caption{ \textbf{Illustration of the \emph{background cheating}}.
	In the real open world, \emph{an} action can happen at \emph{various} locations. Current models trained on the mainstream datasets tend to give predictions simply because it sees some background cues, neglecting the fact that motion pattern is what actually defines an ``action".
	}
	\label{fig:teaser}
\end{figure}

Convolutional neural networks (CNNs) have achieved competitive accuracy on a variety of video understanding tasks, including action recognition \cite{hara2018can}, temporal action detection \cite{zhao2017temporal} and spatio-temporal action localization \cite{weinzaepfel2015learning}.
Such success relies heavily on manually annotated datasets, which are time-consuming and expensive to obtain.
Meanwhile, there are numerous unlabeled data that are instantly available on the Internet, drawing more and more researchers' attention from the community to utilize off-the-shelf unlabeled data to improve the performance of CNNs by self-supervised learning. 

Recently, self-supervised learning methods have been developed from the image field to the video field. However, there are big differences between the mainstream video dataset and the mainstream image dataset.
Li et al.\cite{Li_2018_ECCV} and Girdhar et al.\cite{girdhar2020cater} point out that the current commonly used video datasets usually exist large \emph{implicit biases} over scene and object structure , making the temporal structure become less important and the prediction tends to have a high dependence on the video background. We name this phenomenon as \textit{background cheating}, as is shown in Figure \ref{fig:teaser}. For example, a trained model may classify an action as \emph{playing soccer} simply because it sees the field, without really understanding the \emph{cartwheel} motion. As a result, the model is easily to overfit the training set, and the learned feature representation is likely to be scene-biased. Li et al.\cite{Li_2018_ECCV} reduce the bias by resampling the training set, and Wang et al.\cite{wang2018pulling} propose to pull actions out of the context by training a binary classifier to explicitly distinguish action samples and conjugate samples that are contextually similar to action samples but contains different action.

In this work, to hinder the model from \textit{background cheating} and make the model generalize better, we present to reduce the impact of the background by adding the background and encourage the model to learn consistent feature w/ or w/o the operation. Specifically, given a video, we randomly select a static frame and add it to every other frames to construct a distracting video, as is shown in Figure \ref{fig:spatial_mixup}. Then we force the model to pull the feature of the distracting video and the feature of the original video together by consistency regularization.  
In this way, we made a disturbance to the video background and require its feature to be consistent with the original video, achieving the purpose of making the model not be excessively dependent on the background, thereby alleviating the \textit{background cheating} problem.

Experimental results demonstrate that the proposed method can effectively reduce the influence of the \emph{background cheating}, and the extracted representation is more robust to the background bias and have stronger generalization ability. Our approach is simple and incorporate it into existing self-supervised video learning methods can bring significant gains.
In summary, our main contributions are twofold:
\begin{itemize}
\item We propose a simple yet effective video representation learning method that is robust to the background.
\item The proposed approach can be easily incorporated with existing self-supervised video representation learning methods, bringing further gains on UCF101\cite{soomro2012ucf101}, HMDB51 \cite{kuehne2013hmdb51} and Diving48\cite{li2018resound} datasets.
\end{itemize}
 
\section{Related Work}

\subsection{Self-supervised Learning for Image}
Self-supervised learning is a generic learning framework which gets supervision from the data itself. Current methods can be grouped into two types of paradigms, \emph{i.e.}, constructing pretext tasks or constructing contrastive learning.

\noindent\textbf{Pretext tasks.} These methods focus on solving surrogate classification tasks with surrogate labels, including predicting the rotation angle of image\cite{gidaris2018unsupervised}, solving the jigsaw puzzle\cite{noroozi2016unsupervised}, coloring image\cite{zhang2016colorful} and predicting relative patches\cite{noroozi2016unsupervised}, etc. 
Recently, the type of image transformation also be used as a surrogate\cite{zhang2019aet}.
\noindent\textbf{Contrastive learning.} Another mainstream method is based on contrastive learning, which regards each instance as a category. Early work \cite{dosovitskiy2015discriminative} directly used each sample in the dataset as a category to learn a linear classifier, but this method will become infeasible when the number of samples increases. To alleviate this problem, Wu et al. \cite{wu2018unsupervised} replace the classifier with a memory bank storing previously computed representations and then use a noise contrastive estimation \cite{gutmann2010noise} to compare instances. MoCo \cite{he2020momentum} stores the representations from a momentum encoder and achieves great success. In contrast, Ye et al. \cite{ye2019unsupervised} propose to use a mini batch to replace the memory bank. SimCLR \cite{chen2020simple} shows that the memory bank can be entirely replaced by a large batch size.

\subsection{Self-supervised Video Representation Learning}
Recent years, self-supervised learning has been expanded into the video domain and attracts a lot interests.

\noindent\textbf{Pretext tasks.} The majority of the prior work explore natural video properties as supervision signal. Among them, temporal order is one of the most widely-used property, such as, the arrow of time \cite{wei2018learning}, the order of shuffled frames \cite{misra2016shuffle}, the order of video clip \cite{xu2019self} and the playback rate of the video \cite{benaim2020speednet,yao2020video}.
Besides the temporal order, the spatio-temporal statistics are also used as supervision. 
For example, pxiel-wise geometry information \cite{gan2018geometry}, space-time cubic puzzles \cite{kim2019self,luo2020video} and the optical-flow and the appearance statistics \cite{wang2019self}. In addition, DynamoNet\cite{diba2019dynamonet} predicts future frames by learning dynamic motion filter, which is pre-trained on a large-scale dataset Youtube-8M. More recently, Buchler et al. \cite{buchler2018improving} and ELO \cite{piergiovanni2020evolving} propose to ensemble multiple pretext task based on reinforcement learning.

\noindent\textbf{Contrastive learning.} Contrastive learning is introduced into the field of video representation learning by TCN \cite{sermanet2018time}, which uses different camera views as positive samples. IIC \cite{tao2020selfsupervised} proposes an inter-intra mulit-modal contrastive framework based on the Contrastive Multiview Coding \cite{tian2019contrastive}. CoCLR \cite{han2020self} takes the advantage of the natural correlation between the RGB and the Optical Flow modalities to select the negative samples in the memory bank. GDT \cite{miech2020end} achieves great success by using tens of millions data for pre-training with multi-modal constrastive leanring.

It is worth to mention that while all methods mentioned above focus on designing specific tasks, we present a generalized constraint term that can be integrated into any existing self-supervised video representation learning approach.

\subsection{Background Biases in Video}
Current widely used video datasets have serious bias towards the background\cite{Li_2018_ECCV,girdhar2020cater}, which may misleads the model using just the static cues to achieve good results. For example, only using three frames during training, TSN\cite{wang2018temporal} can achieve 85\% accuracy on UCF101. Therefore, using these datasets for training can easily cause the model making background biased predictions. 

In order to mitigate the background bias, Li et al.\cite{li2018resound} resample the original dataset to generate a less biased dataset Diving48 for the action recognition task.
Wang et al.\cite{wang2018pulling} use conjugate samples that are contextually similar to human action samples but do not contain the action to train a classifier to deliberately separate the action from the context. Choi et al.\cite{choi2019can} propose to detect and mask actors with a human detector and further present a novel adversarial loss for debasing.
In this work, we try to debias through consistency constraint, which is simple but effective and does not need additional costs.

\section{Methodology}

In this section we introduce the proposed \emph{Background Erasing} (BE) method. We first give an overall description of BE, and then introduce how to integrate BE into existing self-supervised methods.

\subsection{Overall Architecture}
\label{sec:overall_archite}

\begin{figure}[hbt!]
	\centering
	\includegraphics[width=0.87\linewidth]{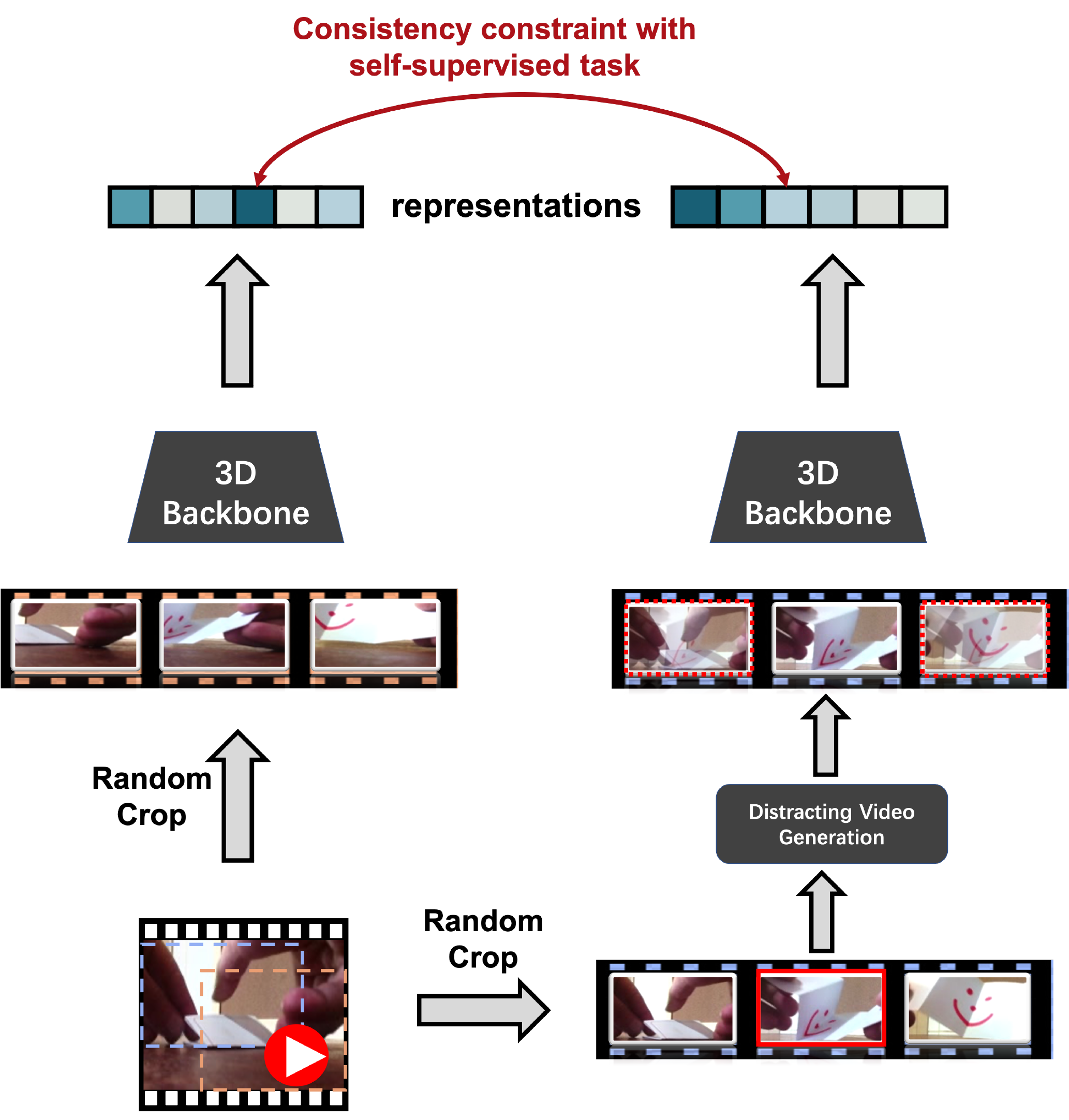}
 	\caption{\textbf{The framework of the proposed method BE.} A video is first randomly cropped spatially, then we generate the distracting video by adding a static frame upon other frames. The model is trained by a existing self-supervised task together with a consistency constraint, with the goal of pulling the feature of the original video and that of the distracting video closer. (Best viewed in color).
	}
	\label{fig:overall}
\end{figure}

The framework of the proposed BE is shown in Figure \ref{fig:overall}. 
For each input video $x$, we first randomly crop two fixed-length clips from different spatial locations, denoted as $x^{o}$ and $x^{v}$. Suppose we have a basic data augmentation set $\mathcal{A}$, from which we sample two specific operations $a^1$ and $a^2$, and operate on $x^{o}$ and $x^{v}$ respectively.
In this way, the input clips have different distribution in the pixel level but are consistent in the semantic level. 
Afterwards, ${x}^{o}$ is directly fed into the 3D backbone to extract the feature representation and we denote this procedure as $F({x}^{o}; \theta)$, where $\theta$ represents the backbone parameters. For $x^{v}$, we first generate a distracting counterpart $x^{d}$ for it, which has the interference of added static frame noise but the semantics remains the same. The output feature maps of ${x}^{o}$ and ${x}^{d}$ are represented by $f_{x^{o}}, f_{x^{d}} \in \mathbb{R}^{C \times T \times H \times W}$. $C$ is the number of channel and $T$ is the length of time dimension. $W$ and $H$ are spatial size. At last, the extracted features $f_{x^{o}}, f_{x^{d}}$ are pulled closer within the existing self-supervised methods.

\subsection{Background Erasing.}
\label{sec:SpatialMixup}

In the video representation learning, sometimes the statistical characteristics of the background will drown out the motion features of the moving subject. Thus it is easy for the model to make predictions based only on the background information. 
And the model is easy to overfit to the training set and has poor generalization on the new dataset.

\begin{figure}
	\centering
	\includegraphics[width=.9\linewidth]{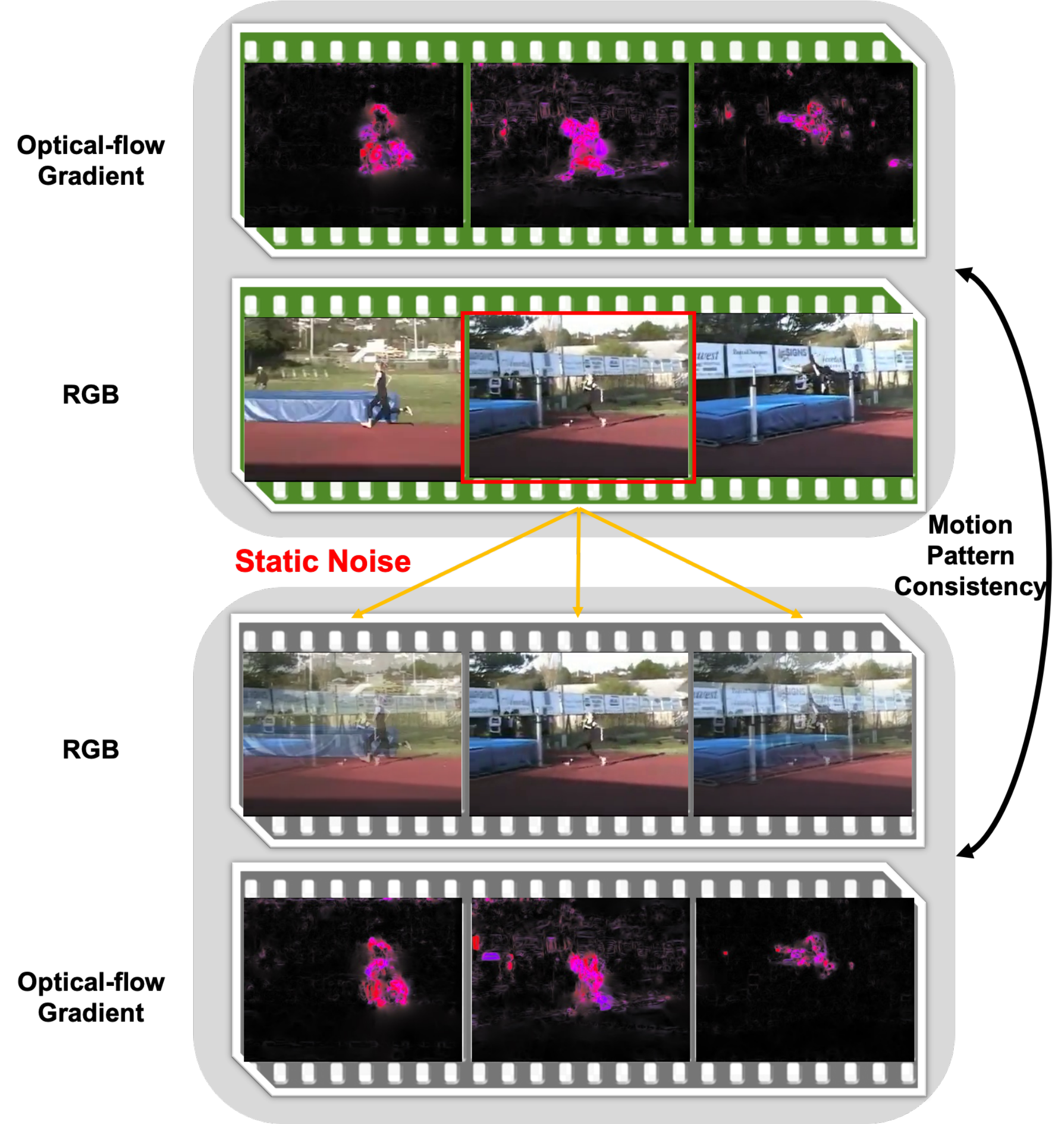}
	\caption{\textbf{Distracting Video Generation}. One intra-video static frame is randomly selected and added to other frames as \textit{Noise}. The background of the generated distracting video has changed, but the optical flow gradient is basically not changed, indicating that the motion pattern is retained.}
	\label{fig:spatial_mixup}
\end{figure}

Background Erasing(BE) is proposed to remove the negative impact of the background by adding the background. Specifically, for a video sequence $x$, we randomly select one static frame and add it as a spatial background noise to every other frames to generate a distracting video, in which each frame $\hat{x}$ is obtained by the following formula:
\begin{equation}
     \hat{x}= (1- \lambda) \cdot x^{(j)} +  \lambda \cdot x^{(k)}, j \in [1,T]
\end{equation}
where $\lambda$ is sampled from the uniform distribution $[ 0, \gamma]$, $x^{(j)}$ means the $j$-th frame of $x$, $k$ denotes the index of the randomly selected frame and $T$ is the length of the video sequence $x$. BE operation is applied to $x^{v}$, and the generated distracting video clip $x^d$ has a background perturbation on the spatial dimension, but the motion pattern is basically not changed, as shown in Figure \ref{fig:spatial_mixup}.

Furthermore, it is easy to prove that the \textit{time derivative} of $x^d$ is a linear transformation of the \textit{time derivative} of $x^{v}$, formally: 
\begin{equation} 
     \frac{\mathrm{d}((1-\lambda)x^{v}+\lambda \delta)}{\mathrm{d}t} = (1-\lambda) \frac{\mathrm{d}x^{v}}{\mathrm{d}t}
\end{equation}
where $\delta$ represents the result of repeating the selected frame $x^{(k)}$ $T$ times along the time dimension. Previous works\cite{1544882,5196739,910878,wang2016temporal} have shown that the \textit{time derivative} of a video clip is an important information for action classification, thus, the property that BE maintains the linear transformation of such information is very crucial.
Afterwards, we force the model to pull the feature of $x^{o}$ and the feature of $x^{d}$ closer, which will be introduced in details later. Since $x^{o}$ and $x^{d}$ resemble each other in the motion pattern but differentiate each other in spatial, when the features of $x^{o}$ and $x^{d}$ are brought closer, the model will be promoted to suppress the background noise, yielding video representations that are more sensitive to motion changes.
We have tried a variety of ways to add background noise, results are shown in Table \ref{tab:mix_cmp}.
Experimental results demonstrate that the intra-video static frame, \emph{i.e.}, BE, works best overall. Meanwhile, we have also tried to add various data augmentations to the selected intra-video static frame to introduce more disturbance, but there is no positive gain.

\subsection{Plug-and-Play}

Using BE solely for optimization will make the model fall into a trivial solution easily. 
Therefore, we integrate BE into the existing self-supervised methods, specifically, we adopt two paradigms, handcrafted pretext and contrastive.

\subsubsection{Pretext Task}
Most pretext tasks can be formulated as a multi-category classification task and optimized with the cross-entropy loss. Specifically, each pretext will define a transformation set $R$ with $M$ operations. Given an input $x$, a transformation $r \in R$ is performed, then the convolutional neural network with parameters $\theta$ is required to distinguish which operation it is. The loss function is as follows:
\begin{equation}
    \mathcal{L}_p =- \frac{1}{M} \sum_{r \in R}^{} \mathcal{L}_{ce}(F(r(x);\theta), r),
\end{equation}
where $\mathcal{L}_{ce}$ is Cross Entropy.

\noindent\textbf{Plugged-in BE.} For handcrafted pretext task, we use a consistency regularization term to pull the feature of $x^{o}$ closer to the feature of $x^{d}$, and make them consistent in the temporal dimension. Formally,
\begin{equation}
    \label{eq:consistencyregularization}
    \mathcal{L}_{be} = {||\psi (f_{ x^{o}}) - \psi (f_{x^{d}})||}^2
\end{equation}
where $\psi$ is an explicit feature mapping function that project features from $C\times T \times H \times W$ to $C \times T$. We use spatial global max pooling since $x^{o}$ and $x^{d}$ have different pixel distribution due to random cropping. In this fashion, we force the max response at each time dimension being consistent. And the final loss is:
\begin{equation}
    \mathcal{L} = \mathcal{L}_p + \beta \mathcal{L}_{be}
\end{equation}
where $\beta$ is a hyperparameter that controls the importance of the regularization term. In our experiments, $\beta$ is set to 1.

\subsubsection{Contrastive Learning}
Contrastive learning \cite{hadsell2006dimensionality} aims to learn an invariant representation for each sample, which is achieved by maximizing similarity of similar pairs over dissimilar pairs.

\noindent\textbf{Plugged-in BE.} Given a video dataset $\mathcal{D}$ with $N$ videos $\mathcal{D}=\{x_1,x_2,...,x_N\}$, for each video $x_i$, we randomly sample once in each epoch, obtaining $x_{i}^{o}$ and $x_{i}^{d}$. In order to add a consistency constraint between $x^o$ and $x^d$, we directly treat their features $f(x^o)$ and $f(x^d)$ as positive pairs instead of using MSE loss. Specifically, assume there is a projection function $\phi$, which consists of a spatio-temporal max pooling and a fully connected layer with $D$ dimension. Then the high level feature can be encoded by $z_x$ = $\phi(f(x))$. Given a particular video $x_i$ and clip sampling function $s$, the negative set $\mathcal{N}_{1i}$ is defined as: $\mathcal{N}_{1i}=\{s(x_n) | \forall n \neq i\}$, each element in $\mathcal{N}_{1i}$ is a clip and represents an identity, then the InfoNCE\cite{oord2018representation} loss is improved as follows:
\begin{equation}
\label{eq:infoNCE}
    \mathcal{L} = -\frac{1}{N}\sum_{i=1}^{N}log\frac{exp(z_{x_{i}^{o}} \cdot z_{x_{i}^{d}})}{exp(z_{x_{i}^{o}} \cdot z_{x_{i}^{d}}) + \sum_{n \in \mathcal{N}_{1i}}^{} exp(z_{x_{i}^{o}}, z_n)}
\end{equation}
where $\cdot$ denotes the dot product. In this way, the optimization goal is \textit{video-level} discrimination in essence.

However, in order to discriminate each instance in $\mathcal{D}$, there may exist many spatial details. In order to make the objective more challenge, we introduce \textit{hard negatives}, the different video clips with augmentation $a^{1}$ but from the same video. In this way, the optimization goal changes from \textit{video-level} into \textit{clip-level}, which is based on the observation that different clips of the same video contain different motion patterns but similar background. The hard negative set $\mathcal{N}_{2i}$ for $x_{i}$ is defined as: $\mathcal{N}_{2i}=\{x_{i}^{h}|x_{i}^{h} \neq x_{i}^{o},x_{i}^{h}\in x_{i} \}$, and the overall negative set is $\mathcal{N}_i = \{\mathcal{N}_{1i} \cup \mathcal{N}_{2i}\}$.
Then the final objective function is:
\begin{equation}
    \label{eq:video_clip}
    \mathcal{L} = -\frac{1}{N} \sum_{i=1}^{N}log\frac{exp(z_{x_{i}^o} \cdot z_{x_{i}^d} )}{exp(z_{x_{i}^{o}} \cdot z_{x_{i}^{d}}) + \sum_{n \in \mathcal{N}_{i}}^{} exp(z_{x_{i}^{o}} \cdot z_n)}
\end{equation}
For efficiency, we randomly select one hard negative sample from $\mathcal{N}_{2i}$ each iteration and we find more hard negative samples have a similar result experimentally.

\begin{table*}
    \centering
    {
    \footnotesize
    \begin{tabular}{lll|lllcl|ll}
    \shline
    \multicolumn{3}{c}{\textbf{Method}}&\multicolumn{5}{c}{\textbf{Pretrain}}&\multicolumn{2}{c}{\textbf{Fine-tune}}\\
    \hline
    {\bf Method(year)}&{\bf Backbone}&{\bf Depth}&{\bf Dataset(duration)}&{\bf Frame}&{\bf Res}&{\bf Single-Mod}&{\bf \textit{C}/\textit{P}}& {\bf UCF101} & {\bf HMDB51} \\
    \hline
    \multicolumn{1}{l}{\textbf{Supervised}}&&&&&&&&\\
    Random Init &I3D&22&\redxmark&-&224&\greencmark&-& 60.5 & 21.2 \\
    ImageNet Supervised &I3D&22&ImageNet&-&224 &\greencmark&-&67.1 & 28.5 \\
    K400 Supervised &I3D&22&K400(28d)&-&224&\greencmark&-&96.8 & 74.5\\
    \hline
    \multicolumn{1}{l}{\textbf{Self-supervised}}&&&&&&&\\
    Shuffle \cite{misra2016shuffle}  \textcolor[rgb]{0,0,1}{[ECCV, 2016]} &AlexNet&8&UCF101(1d)&-&112&\greencmark&\textit{P}&50.2 & 18.1 \\
    VGAN \cite{vondrick2016generating}  \textcolor[rgb]{0,0,1}{[NeurlPS, 2016]} &VGAN&22&UCF101(1d)&-&112&\greencmark&\textit{P}& 52.1 & - \\
    OPN \cite{lee2017unsupervised}  \textcolor[rgb]{0,0,1}{[ICCV, 2017]} &Caffe Net&14&UCF101(1d)&-&112&\greencmark&\textit{P}& 56.3 & 22.1 \\
    Geometry \cite{gan2018geometry} \textcolor[rgb]{0,0,1}{[CVPR, 2018]} &Flow Net&56&UCF101(1d)&16&112&\redxmark&\textit{P}& 55.1 & 23.3\\
    IIC \cite{tao2020selfsupervised} \textcolor[rgb]{0,0,1}{[ACM MM, 2020]} &C3D &10& UCF101(1d)&16&112&\redxmark&\textit{C}&72.7&36.8\\
    Pace \cite{wang2019self} \textcolor[rgb]{0,0,1}{[ECCV, 2020]}&R(2+1)D&23&K400(28d)&16&112&\greencmark&\textit{C}&77.1&36.6\\
    \hdashline
    3D RotNet \cite{jing2018self} \textcolor[rgb]{0,0,1}{[2018]} & C3D &10 & K400(28d)&16&112&\greencmark&\textit{P}&62.9 & 33.7\\
    \textbf{3D RotNet + BE} &C3D&10&K400(28d)&16&112&\greencmark&\textit{P}& \textbf{65.4}\greenp{2.5$\uparrow$} & \textbf{37.4}\greenp{3.7$\uparrow$} \\
    \hdashline
    ST Puzzles \cite{kim2019self}  \textcolor[rgb]{0,0,1}{[AAAI, 2019]} &C3D&10&UCF101(1d)&48&112&\greencmark&\textit{P}& 60.6 & 28.3\\
    \textbf{ST Puzzles + BE} &C3D&10&UCF101(1d)&48&112&\greencmark&\textit{P}& \textbf{63.7}\greenp{3.1$\uparrow$} & \textbf{30.8}\greenp{2.5$\uparrow$}\\
    \hdashline
    Clip Order \cite{xu2019self}  \textcolor[rgb]{0,0,1}{[CVPR, 2019]} &C3D&10&UCF101(1d)&64&112&\greencmark&\textit{P}& 65.6& 28.4\\
    \textbf{Clip Order + BE}&C3D&10&UCF101(1d)&64&112&\greencmark&\textit{P}&\textbf{68.5}\greenp{2.9$\uparrow$} & \textbf{32.8}\greenp{4.4$\uparrow$}  \\
    \hdashline
    MoCo \cite{he2020momentum} \textcolor[rgb]{0,0,1}{[CVPR, 2020]}$\Diamond$& C3D & 10 & UCF101(1d)&16&112&\greencmark&\textit{C}&60.5&27.2 \\
    \textbf{MoCo + BE}  & C3D & 10 & UCF101(1d)&16&112&\greencmark&\textit{C}& \textbf{72.4}\greenp{11.9$\uparrow$}&\textbf{42.3}\greenp{14.1$\uparrow$} \\
    \hdashline
    CoCLR\cite{han2020self} \textcolor[rgb]{0,0,1}{[NeuIPS, 2020]} &R3D&23&K400(28d)&32&128&\redxmark&\textit{C}&87.9&54.6\\
    DPC~\cite{han2019video}\textcolor[rgb]{0,0,1}{[ICCW, 2019]} &R3D&34&K400(28d) & 64&224&\greencmark&\textit{P} & 75.7 & 35.7 \\
    AoT~\cite{wei2018learning} \textcolor[rgb]{0,0,1}{[CVPR, 2018]} &T-CAM &-&K400(28d)&64&224&\greencmark&\textit{P} & 79.4 & -   \\
    Pace \cite{wang2019self} \textcolor[rgb]{0,0,1}{[ECCV, 2020]}&S3D-G&23&K400(28d)&64&224&\greencmark&\textit{C}&87.1&52.6\\
    SpeedNet \cite{benaim2020speednet} \textcolor[rgb]{0,0,1}{[CVPR, 2020]} &S3D-G&23&K400(28d)&64&224&\greencmark&\textit{P}&81.1&48.8\\
    SpeedNet \cite{benaim2020speednet} \textcolor[rgb]{0,0,1}{[CVPR, 2020]} &I3D&22&K400(28d)&64&224&\greencmark&\textit{P}&66.7&43.7\\
    \hdashline
    MoCo \cite{he2020momentum} \textcolor[rgb]{0,0,1}{[CVPR, 2020]}$\Diamond$&I3D&22&K400(28d)&16&224&\greencmark&\textit{C}&70.4&36.3\\
    {\bf MoCo + BE}&I3D&22&K400(28d)&16&224&\greencmark&\textit{C}&\textbf{86.8}\greenp{16.4$\uparrow$}&\textbf{55.4}\greenp{19.1$\uparrow$}\\
    \hdashline
    {\bf MoCo + BE}&I3D&22&UCF101(1d)&16&224&\greencmark&\textit{C}&82.4&52.9\\
    {\bf MoCo + BE}&R3D&34&UCF101(1d)&16&224&\greencmark&\textit{C}&83.4&53.7\\
    {\bf MoCo + BE}&R3D&34&K400(28d)&16&224&\greencmark&\textit{C}&87.1&56.2\\
    \shline
    \end{tabular}
    }
    \caption{Top-1 accuracy (\%) of integrating BE as a regularization term to four existing approaches and compared with previous methods on the UCF101 and HMDB51 dataset. Single-Mod denotes Single-Modality, C/P represents Contrastive/Pretext task, $\Diamond$ means our implementation, K400 is short for Kinetics-400 and d represents day.}
    \label{tab:sota_action_recognition_cmp}
\end{table*}
\section{Experiments}

\subsection{Implementation Details}
\label{sec:dataset_and_evaluations}

\noindent\textbf{Datasets.} All the experiments are conducted on four 
video datasets, UCF101 \cite{soomro2012ucf101}, HMDB51 \cite{kuehne2013hmdb51}, Kinetics \cite{kay2017kinetics} and Diving48 \cite{li2018resound}. The first three contain prominent bias, while Diving48 is less biased. UCF101 is a realistic video dataset with 13,320 videos of 101 action categories. 
HMDB51 contains 6,849 clips of 51 action categories. Kinetics is a large scale action recognition dataset that contains 246k/20k train/val video clips of 400 classes. Diving48 consists of ~18k trimmed video clips of 48 diving sequences. 

\noindent\textbf{Networks.} We use C3D \cite{tran2015learning}, R3D \cite{hara2018can} and I3D \cite{carreira2017quo} as base encoders followed by a spatio-temporal max pooling layer.

\noindent\textbf{Default Settings.} All the experiments are conducted on 8 Tesla V100 GPUs with a batch size of 64 under PyTorch\cite{paszke2017automatic} framework. We adopt SGD as our optimizer with momentum of 0.9 and weight decay of 5e-4. 

\noindent\textbf{Self-supervised Pre-training Settings.}
We pre-train the network for 50 epochs with the learning rate initialized as 0.01 and decreased to 1/10 every 10 epochs. The input clip consists of 16 frames, which is uniformly sampled from the original video with a temporal stride of 4. Then the sampled clip is resized to $16 \times 3 \times 112 \times 112$ or $16 \times 3 \times 224 \times 224$. The $\gamma$ of Background Erasing is experimentally set to 0.3, and a larger value may result in excessive blur. The choice of temporal stride and $\gamma$ is analysed in the supplementary. The basic augmentation set $\mathcal{A}$ contains random rotation less than 10 degrees and color jittering, and all these operations are applied in a temporal consistent way, that is, each frame of a video uses the same augmentation. The vector dimension $D$ is 128.

\noindent\textbf{Supervised Fine-tuning Settings.}
After pre-training, we transfer the weights of the base encoder to two downstream tasks, \emph{i.e.}, action recognition and video retrieval, with the last fully connected layer randomly initialized. We fine-tune the network for 45 epochs. The learning rate is initialized as 0.05 and decreases to 1/10 every 10 epochs. 

\noindent\textbf{Evaluation Settings.} For action recognition, following common practice\cite{xu2019self}, the final result of a video is the average of the results of 10 clips that are uniformly sampled from it during testing time.

\begin{table} [t]
\centering
\resizebox{\columnwidth}{!}{%
{\footnotesize
		\begin{tabular}{l|lc|l} 
		\shline
		\multicolumn{1}{l}{\textbf{Method}}&\multicolumn{1}{l}{\textbf{Pretrain}}&\multicolumn{1}{c}{\textbf{Single-Mod}}&\multicolumn{1}{l}{\textbf{Diving48}}\\
			 \hline 
			{\textbf{Supervised Learning}}&&&\\
			R(2+1)D \cite{tran2018closer}\textcolor[rgb]{0,0,1}{[CVPR, 2018]} & \redxmark & \greencmark & 21.4\\
 			R(2+1)D \cite{tran2018closer} \textcolor[rgb]{0,0,1}{[CVPR, 2018]}& Sports1M & \greencmark & 28.9	 \\
 			I3D\cite{carreira2017quo}$\Diamond$\textcolor[rgb]{0,0,1}{[CVPR, 20187]}&ImageNet&\greencmark&20.5\\
			I3D\cite{carreira2017quo}$\Diamond$\textcolor[rgb]{0,0,1}{[CVPR, 2017]}&K400&\greencmark&27.4\\
 			TRN~ \cite{zhou2018temporal} \textcolor[rgb]{0,0,1}{[ECCV, 2018]} & ImageNet  & \redxmark & 22.8 \\
 			DIMOFS \cite{bertasius2018learning} \textcolor[rgb]{0,0,1}{[2018]}& K400+Track & \redxmark & 31.4 \\
 			GST \cite{luo2019grouped} \textcolor[rgb]{0,0,1}{[ICCV, 2019]}& ImageNet & \greencmark & 38.8\\
 			Att-LSTM \cite{kanojia2019attentive} \textcolor[rgb]{0,0,1}{[CVPRW, 2019]}&  ImageNet & \greencmark & 35.6 \\
 			GSM \cite{sudhakaran2020gate} \textcolor[rgb]{0,0,1}{[CVPR, 2020]}&ImageNet&\greencmark&40.3\\
            CorrNet \cite{wang2020video} \textcolor[rgb]{0,0,1}{[CVPR, 2020]}& Sports1M & \greencmark  & 44.7 \\  
 \hline 
 {\textbf{Self-supervised Learning}}&&&\\
   {\bf MoCo + BE (I3D)}  & Diving48 &\greencmark &  58.3 \\
    \hdashline 
    {\bf MoCo + BE (R3D-18)}  & UCF101 &\greencmark &  46.6\\
 \hdashline 
  MoCo \cite{he2020momentum} $\Diamond$ (I3D)  & UCF101 & \greencmark & 43.2 \\
 {\bf MoCo + BE (I3D)}  & UCF101 &\greencmark &  {\bf 58.8}\greenp{15.6$\uparrow$}\\
 \hdashline
 MoCo \cite{he2020momentum} $\Diamond$ (I3D) & K400 & \greencmark & 47.9\\
 {\bf MoCo + BE (I3D)}  & K400 &\greencmark & {\bf 62.4}\greenp{14.5$\uparrow$}\\
  \hline
		\end{tabular}
}}
	\caption{Top-1 accuracy (\%) of integrating BE into MoCo and compared to previous method on Diving48. 
	}
	\label{tab:soa_some_dive}
\end{table}

\subsection{Action Recognition}
\noindent\textbf{Comparison on common datasets.} In this section, we integrate BE into three pretext tasks, \emph{i.e.}, 3D RotNet, ST Puzzles and Clip Order, and one contrastive task, \emph{i.e.}, MoCo\cite{he2020momentum}, to verify the performance gains brought by BE.
All the results shown in Table \ref{tab:sota_action_recognition_cmp} are averaged over 3 dataset splits. We also report the result of the random initialized model and the result of the model pre-trained with all labels of ImageNet and Kinetics in a supervised manner for reference. It can be observed that plugging BE into three handcrafted pretext tasks can all bring improvements. Specifically, BE brings 2.5\%/3.7\% improvement with 3D RotNet, 3.1\%/2.5\% gain with ST Puzzle and 2.9\%/4.4\% improvement with Clip Order on UCF101/HMDB51. Further, when BE is introduced into MoCo, using the same backbone I3D and the same pretrain dataset Kinetics, it can bring 16.4\% and 19.1\% improvements on UCF101 and HMDB51 respectively, which is significant and nonnegligible. 

\noindent\textbf{Comparison on a less biased dataset.} In this section, we fine-tune and test on a less biased Diving48, and the results are shown in Table \ref{tab:soa_some_dive}. It can be observed that without using additional videos during pre-training, \emph{i.e.}, pre-training and fine-tuning both on Diving48, MoCo enhanced with BE can achieve 58.3\% top-1 accuracy using I3D backbone, which is far beyond the result of Kinetics supervised pre-training (27.4\%). 
When Kinetics is also used in a self-supervised manner, the accuracy of our method can be further improved from 58.3\% to 62.4\%, which achieves state-of-the-art. It proves that our method can well alleviate the negative impact of scene bias in the training set, prevent the model from overfitting to the training set, \emph{i.e.}, hinder the model from background cheating and obtain a more robust representation towards the motion. At the same time, it also indicates that given a dataset with less bias, the benefit from supervised pre-training on a large biased dataset is very small.

\begin{table}[t]
\centering
\footnotesize
\begin{tabular}{p{2.2cm}p{1cm}|p{0.4cm}p{0.4cm}p{0.4cm}p{0.4cm}p{0.4cm}}
\toprule
\bf Method & \bf Net & \bf 1 & \bf 5 & \bf 10  & \bf 20 & \bf 50\\
\midrule
Clip Order~\cite{xu2019self} & C3D & 7.4 & 22.6 & 34.4 & 48.5& 70.1 \\
Clip Order~\cite{xu2019self} & R3D & 7.6 & 22.9 & 34.4 &48.8 & 68.9 \\
VCP \cite{luo2020video} & C3D& 7.8 & 23.8 & 35.3 & 49.3 &71.6 \\
MemDPC \cite{han2020memory} &R3D&7.7&25.7&40.6&57.7&- \\
Pace \cite{wang2020self} & R3D&9.6 &26.9& 41.1& 56.1& 76.5\\
MoCo \cite{he2020momentum} $\Diamond$ & C3D & 9.5&25.4&38.3&52.2&72.4 \\
\midrule
\textbf{MoCo + BE} & C3D & 10.2&27.6&40.5&56.2&76.6 \\
\textbf{MoCo + BE} & I3D & 9.3 & 28.8 & 41.4 & 57.9 & 78.5 \\
\textbf{MoCo + BE} & R3D & \textbf{11.9} & \textbf{31.3} & \textbf{44.5} & \textbf{60.5} & \textbf{81.4} \\
\bottomrule
\end{tabular}
\caption{{\bf Recall-at-topK (\%).} Accuracy under different K values on HMDB51. 
}
\label{tab:recallatk_hmdb51}
\end{table}

\subsection{Video Retrieval}
In this section, we evaluate BE on video retrieval tasks. Following the convention \cite{wang2019self, benaim2020speednet}, the network is fixed as a feature extractor after pre-training on the split 1 of UCF101. Then the videos from HMDB51 are divided into clips in units of 16 frames. All the clips in the training set constitute a \textit{Gallery}, and each clip in the test set is used as a \textit{query} to retrieve the most similar clip in the \textit{Gallery} with cosine distance. If the category of the query appears in the K-nearest neighbors retrieved, then it is considered as a hit. It should be noted that in order to keep the scale of representations generated by each 3D architecture consistent, we replaced the original global average pooling with an adaptive max pooling, yielding representations with a fixed scale of $1024 \times 2 \times 7 \times 7$. We show the accuracy when $K = 1, 5, 10, 20, 50$ and compare with other self-supervised methods on  HMDB51 in Table \ref{tab:recallatk_hmdb51}. It can be seen that when using the backbone C3D, combining BE with MoCo can bring a 0.7\% improvement to top1 acc and a 2.2\% improvement to top5 acc, which significantly exceeds the Clip Order and VCP with the same backbone. In addition, when using R3D as the backbone, our results surpass the current mainstream method Pace, which proves that the extracted representations are more discriminative.

\subsection{Variants of Distracting Video Generation}
In this section, we conduct experiments to explore the effectiveness of different distracting video generation methods. 
We employ MoCo with I3D as the baseline and optimized with Eq. \ref{eq:video_clip}, all the experiments are pre-trained on the split 1 of the UCF101. 

One main operation in the background erasing is to generate a distracting video with background noise while retaining the temporal semantics.
In order to explore whether adding a static frame is the most effective operation, we compare it with another four common ways: (a).Gaussian Noise: add an identical White Gaussian Noise on each frame. (b).Video Mixup \cite{zhang2018mixup}: interpolate two videos frame by frame. (c).Video CutMix \cite{yun2019cutmix}: randomly replace one region of each frame with a patch from another frame. (d).Inter-Video Frame: randomly select one frame from another video, and add this static frame as noise to each frame of this video. (e).Our Intra-Video Frame: randomly select one frame from the video itself, and add this static frame as noise to each frame of this video. The results are shown in Table \ref{tab:mix_cmp} and three observations can be obtained:

\begin{table} [t]
\centering
{
\footnotesize
\begin{tabular}{l|ll}
\shline
\multicolumn{1}{l}{\textbf{Method}}&\multicolumn{1}{l}{\textbf{UCF101}}&\multicolumn{1}{l}{\textbf{HMDB51}}\\
\hline
baseline & 72.7 &42.1 \\
Gaussian Noise & 73.2\greenp{0.5$\uparrow$}&42.4\greenp{0.3$\uparrow$} \\
Video Mixup &  68.3\redp{4.4$\downarrow$} & 38.1\redp{4.0$\downarrow$} \\
Video CutMix & 71.2\redp{1.5$\downarrow$} & 40.5\redp{1.6$\downarrow$} \\
Inter-Video Frame &  77.4\greenp{4.7$\uparrow$}& 46.5\greenp{4.4$\uparrow$} \\
{\bf Intra-Video Frame} &\textbf{82.4}\greenp{9.7$\uparrow$}&\textbf{52.9} \greenp{10.8$\uparrow$} \\

\shline
\end{tabular}
}
\caption{Top-1 accuracy (\%) of different distracting video generation methods on UCF101 and HMDB51.}
\label{tab:mix_cmp}
\end{table}

\begin{figure}[b]
	\centering
    \includegraphics[width=.75\linewidth]{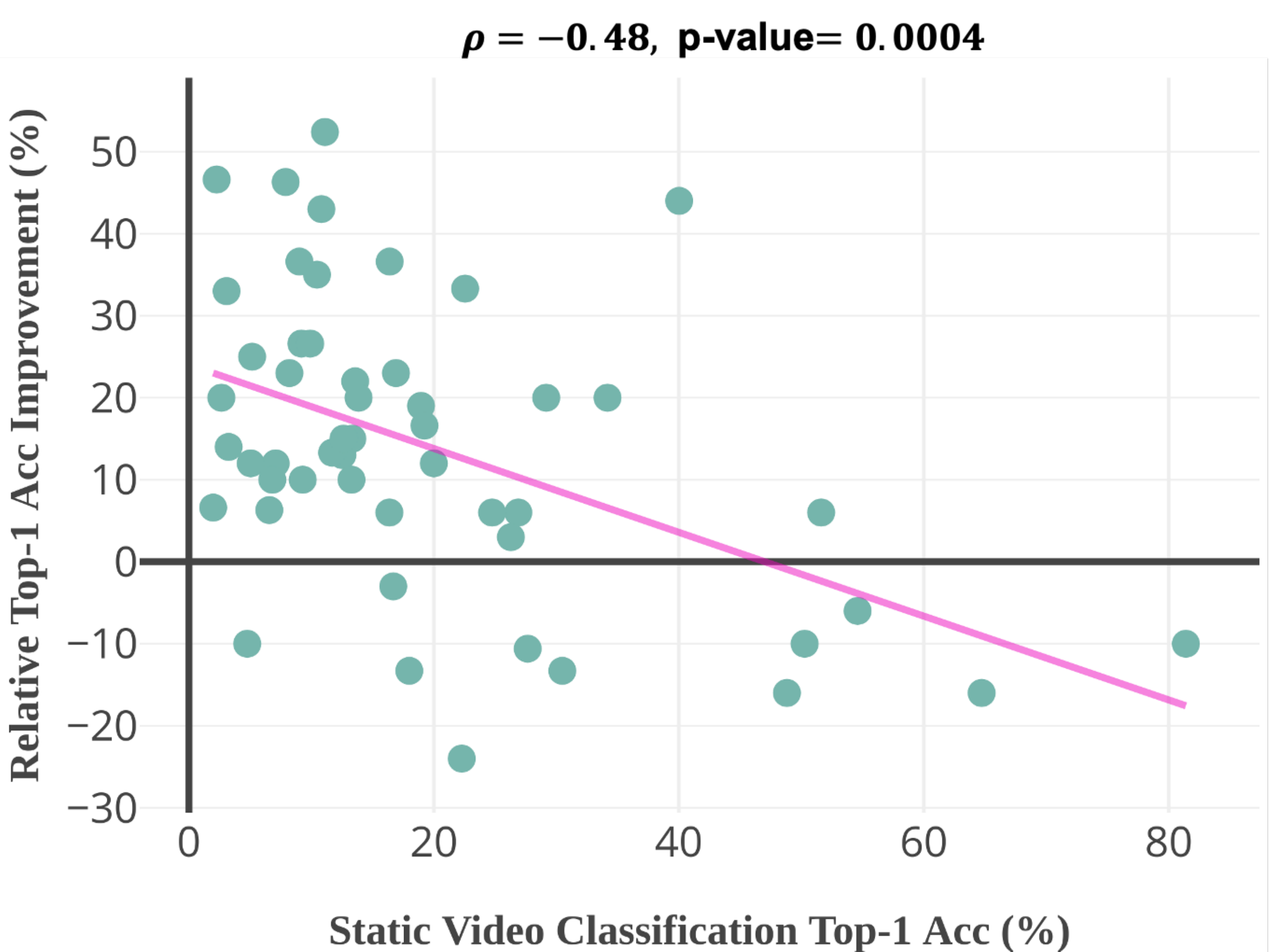}  
	\caption{\textbf{Relative top-1 acc improvement has a strong negative correlation with the static video classification top-1 acc .} Each dot represents a class in HMDB51 dataset and BE brings more significant improvements in categories that rely less on static information.
	}
	\label{fig:relative_bias}
\end{figure}

\emph{i}. Video Mixup and Video CutMix perform worse than the baseline. Notice that these two ways destroy the motion pattern of the original video, which demonstrates the importance of keeping semantics consistency. 

\emph{ii}. Gaussian Noise, Inter-Video Frame and Intra-Video Frame give positive improvement and are more suitable for action modeling since all of them preserve the motion semantics. Therefore, the idea of removing noise by adding noise is effective, but it is essential to make sure the introduced noise does not affect the motion pattern. 

\emph{iii}. Interestingly, we find that Intra-Video Frame leads to 5\% and 6.4\% improvement on UCF101 and HMDB51 respectively compared to the Inter-Video Frame. The only difference between them is the source of the static frame, \emph{i.e.}, the former one is selected from the same video that has a more similar background while the latter one is selected from another video that has more discrepancy. Generally, the background in the video is basically unchanged relative to the motion area. Compared to inter-frame, the scene information added by the intra-frame has the same pixel distribution as most other frames in the video. When the convolutional neural network pulls the feature of the distracting video and that of the original video closer, the model needs to remove static intra-frame noise, which is equivalent to remove the background information in the video, making the extracted feature more robust to the background bias. 
\begin{figure}[t]
  \centering
  \includegraphics[width=\linewidth]{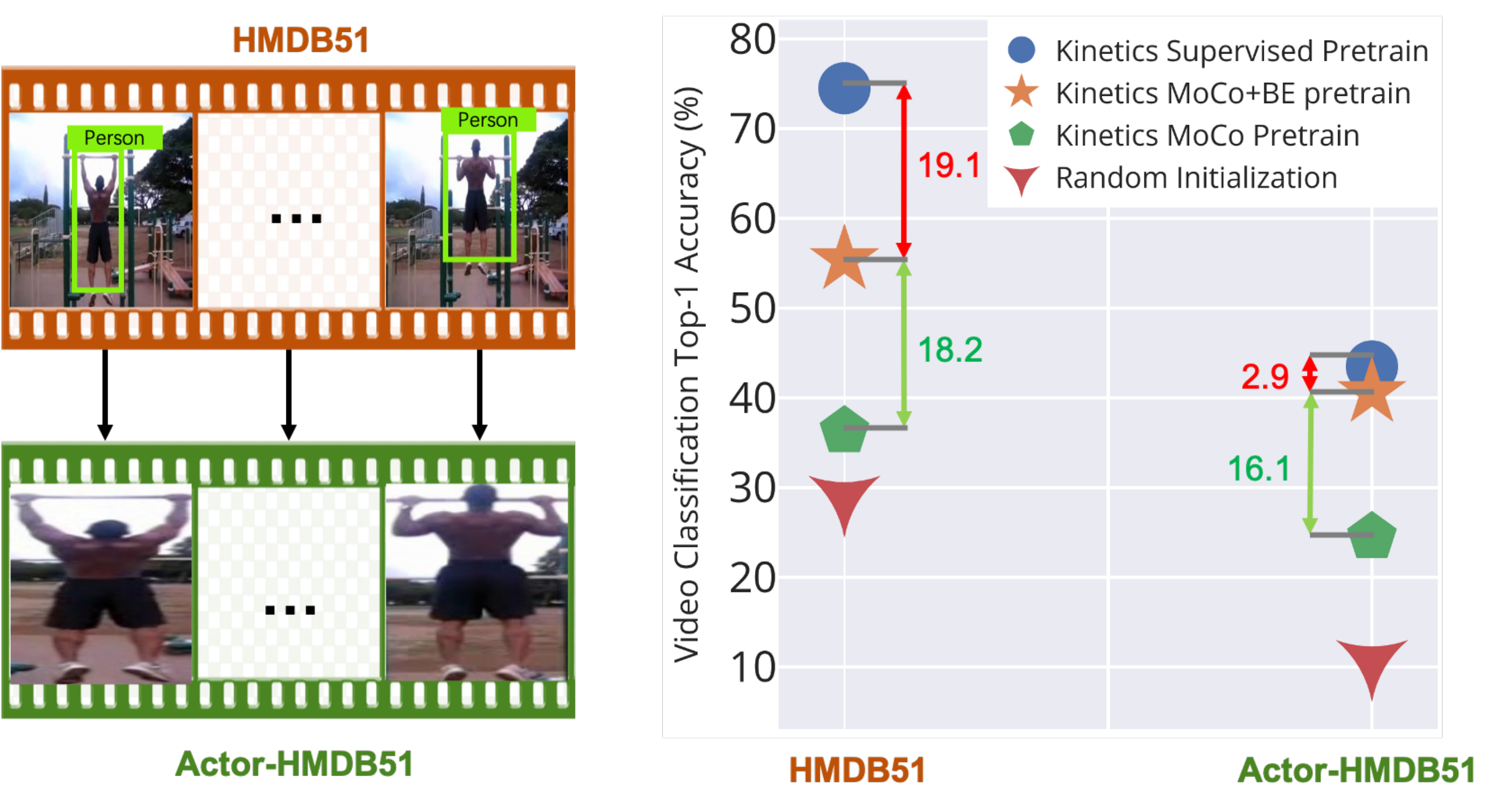}
  \caption{
  Fine-tuning on the actor dominated dataset actor-HMDB51, our method is very close to the result of Kinetics fully supervised, with only 2.9\% difference. Meanwhile the improvement brought by BE over MoCo baseline has only a small drop compared to HMDB51, from 18.2\% to 16.1\%.}
  \label{fig:background_remove}
\end{figure}

\subsection{How does Background Erasing Work?}
In this section, we explore how does the Background Erasing works. To this end, we study the relationship between relative performance improvement (\%) from the proposed Background Erasing and static video classification top-1 accuracy (\%) to see which classes benefit more from our method. \textit{Static} video is generated by randomly selecting one frame and then copying it multiple times. We first trained a randomly initialized I3D model with \textit{static} video generated from HMDB51, which means only \textit{static} information is used. Then two I3D models are pretrained on Kinetics and fine-tuned on HMBD51 using MoCo w/ or w/o BE. At last, we calculate the relative performance improvement brought by BE \emph{w.r.t.} the MoCo baseline, as is shown in Figure \ref{fig:relative_bias}. The Pearson Correlation is $\rho=-0.48$ with a $p$-value 0.0004, indicating a \textit{strong negative correlation} between relative performance improvement and static scene bias. Thus, BE works by bringing a significant improvement in categories that rely less on static information.

\begin{figure}[h]
  \centering
  \includegraphics[width=\linewidth]{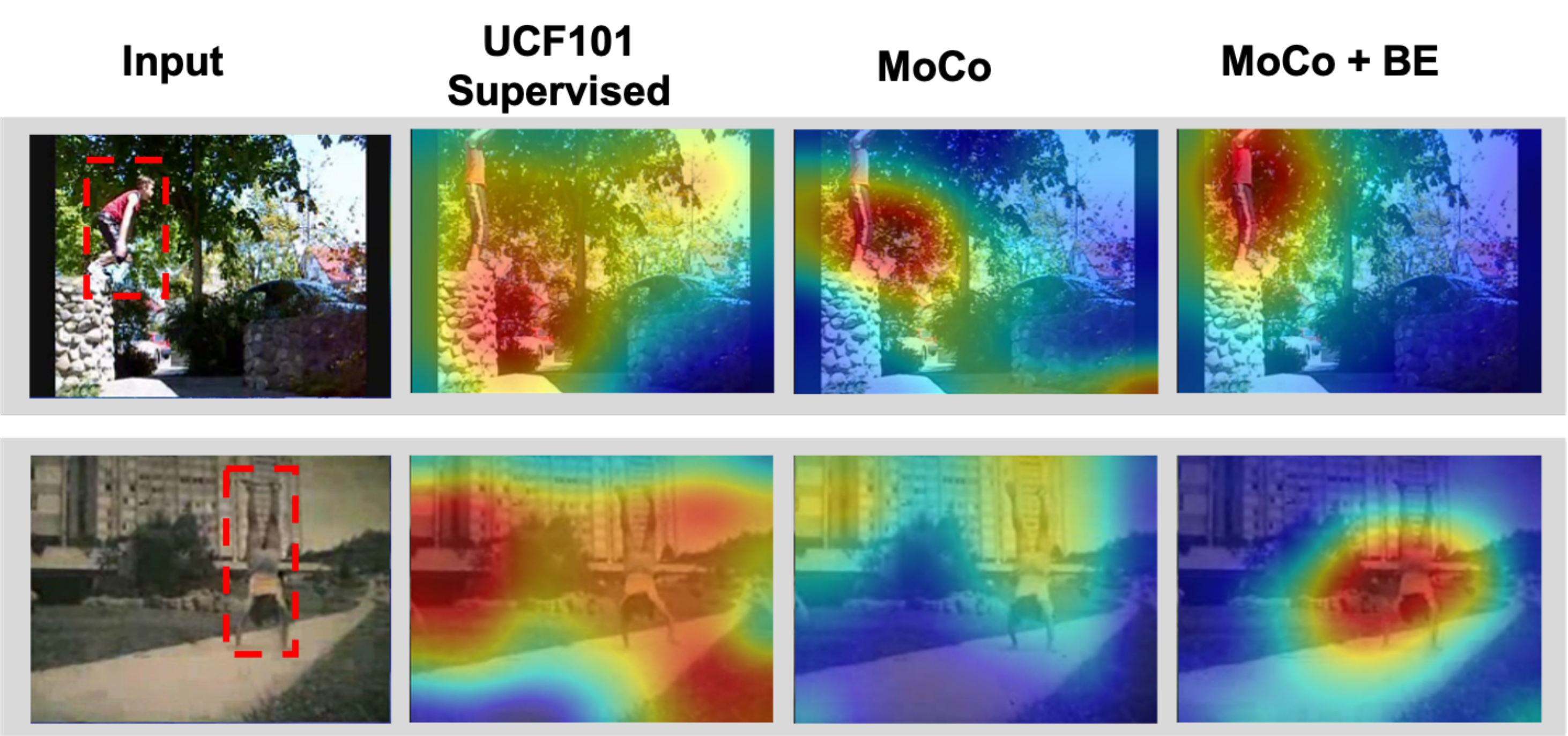}
  \caption{\textbf{Generalization ability on novel classes}. Supervised model is severely affected by the scene bias, while after pre-training with MoCo+BE, the model can precisely focus more on moving areas.}
  \label{fig:transfer}
\end{figure}

\subsection{Is Background Really Removed?}

In order to verify whether BE has really achieved the purpose of removing the background and paying more attention to the moving subject, we tried to cut the background in the real dataset to test the robustness of our work. We first use HMDB51 to generate an actor dominated dataset Actor-HMDB51. Specifically, we detect the actor in each video frame by frame with public implementation \footnote{https://github.com/endernewton/tf-faster-rcnn} of Faster R-CNN \cite{ren2015faster} and then crop actor regions out. Considering that some actions in HMDB51 contain two or more persons, we crop a minimum area that contains all persons for these cases. The dataset Actor-HMDB51 obtained in this way has small scene bias thus requires more attention towards the motion information to be well distinguished. Then, we select biased Kinetics for supervised and self-supervised pre-training, and fine-tune on small scale Actor-HMDB51 using I3D backbone. Figure \ref{fig:background_remove} illustrates the result of different methods on HMDB51 and actor-HMDB51. The performance gap between supervised pre-training and self-supervised MoCo+BE on HMDB51 is 19.1\%(74.5\%-55.4\%), while on Actor-HMDB51 is only 2.9\% (43.5\%-40.6\%), which manifests that the advantages of supervised pre-training heavily rely on background information. However, the improvement brought by BE over the MoCo baseline only slightly decreased, from 18.2\% to 16.1\%. This phenomenon indicates regardless of whether the fine-tuning and test dataset have significant scene bias, BE can steadily bring significant improvement, which demonstrates that BE can indeed make the model pay more attention to the motion pattern. More details about the generation and evaluation of Actor-HMDB51 are provided in the supplementary.

\begin{figure}[h]
  \centering
  \includegraphics[width=\linewidth]{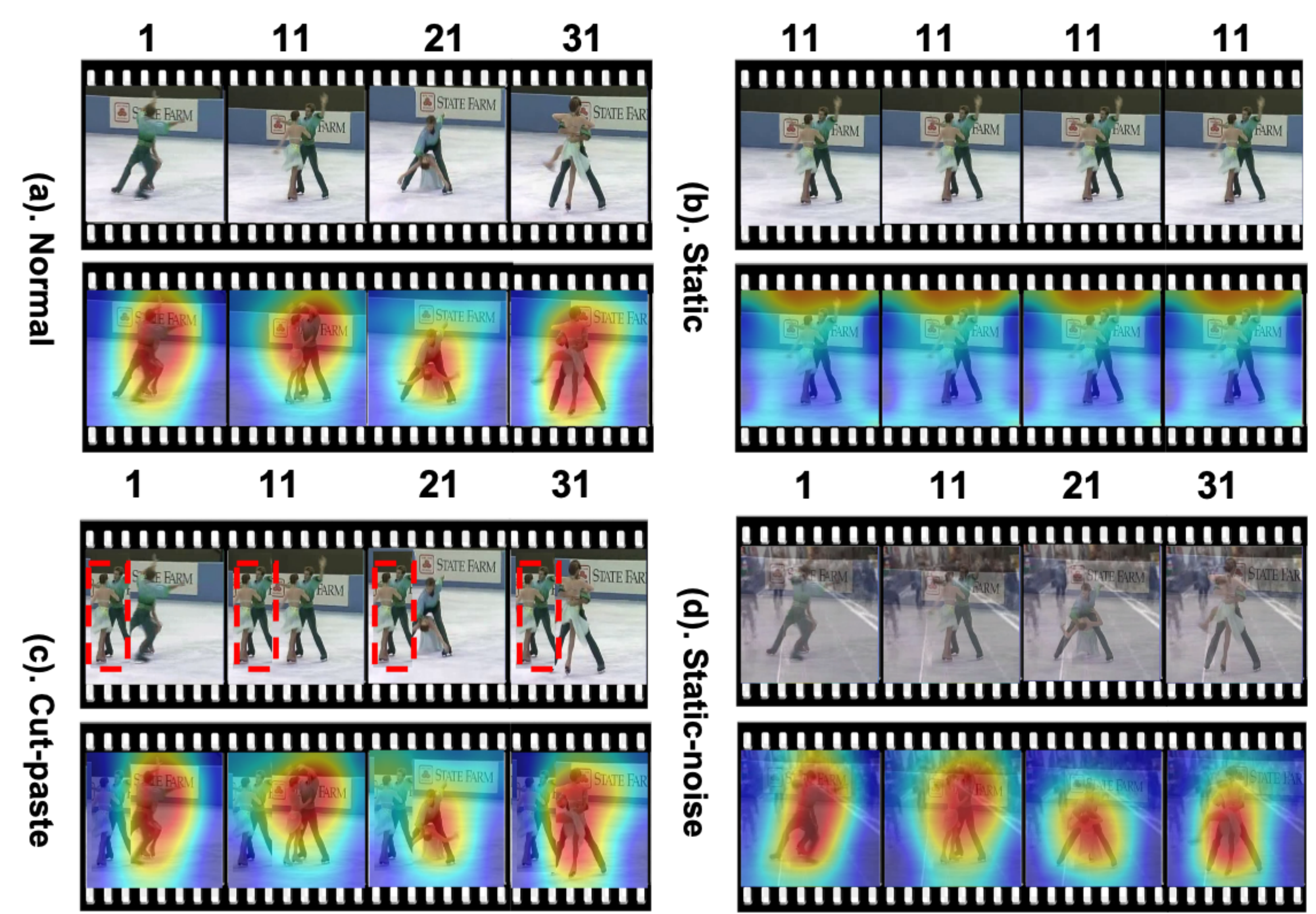}
  \caption{\textbf{Does the model really learn to focus on motion pattern?} \emph{i.} When the input is a \textit{static} video, our model doesn't show high activation as expected. \emph{ii.} When pasting a static human body, the model still focus on the moving persons. \emph{iii.} Using a static frame as background noise does not affect the model focus.}
  \label{fig:adversarial_attack}
\end{figure}

\subsection{Visualization Analysis}
In this part, we visualize the salient regions of the extracted representations with small modifications on CAM\cite{zhou2016learning}. Specially, we select some videos with significant movements of shape $3 \times 16 \times 224 \times 224$ and the extracted feature representations before global average pooling layer is of shape $512 \times  4 \times 4 \times 4$. Then we average these features over the channel dimension to get the compressed features of shape $4 \times 4 \times 4$. Afterwards, the compressed features are resized to the size of original videos and masked to them.

\label{sec:deeper_intra_sample_mixup}
\noindent\textbf{Novel class transfer capability.} To verify the transfer ability of our model on novel class, we visualize some new classes that have never been seen during the training procedure. Specifically, we train three I3D models on UCF101 in both supervised and self-supervised (MoCo and MoCo+BE) manner, and further evaluate on another dataset HMDB51. The visualizations are shown in Figure \ref{fig:transfer}. It can be observed that the supervised model is severely affected by the scene bias and falsely focus on the static background. On the contrary, that the model focus more on motion areas after pre-training with BE and suffer less from scene bias.

\noindent\textbf{Adversarial samples.} In this part, we construct some adversarial samples to verify whether our model can really focus on motion pattern, as shown in Figure \ref{fig:adversarial_attack}. We use MoCo combined with BE, with I3D as backbone and Kinetics as pretrain dataset. First, using a \textit{static} video as input, our model has a low response to the overall area. Then we paste another static human body, our method can correctly focus on the moving actor, which indicates that our model does not only focus on the human body. In addition, we introduce a static frame from \textit{ride bike} action as noise, which will not affect our model. These experiments prove that feature representations extracted by our method have a fully understanding of space-time.

\section{Conclusion}
In this paper, we propose a novel Background Erasing (BE) method for self-supervised learning. 
The proposed method minimizes the feature distance between the sample and sample variation constructed by BE. 
The proposed method is evaluated using different CNN backbones on three benchmark datasets. 
Experimental results show that the proposed BE can be well integrated into both the \emph{pretext task} and \emph{contrastive learning} and outperforms existing methods for action recognition notably, especially on a less biased dataset.

{\small
\bibliographystyle{ieee_fullname}
\bibliography{egbib}
}

\end{document}